# The comparison of density estimation for transfer learning in Banach, Hilbert, Euclidean spaces

Maryam Alsadat Kazemian[1], Azam Bastanfard[1*]
[1] Department of Computer Engineering Ka.C., Islamic Azad University
, Karaj, Iran
* Corresponding author: Azam Bastanfard: [bastanfard@kiau.ac.ir]

## Abstract

Domain adaptation or DA methods are indispensable when target labels or computational resources are scarce, yet the domain kernel density and subspace alignment approaches completely assume a Hilbertian geometry.

They measure distributional discrepancy with symmetric $\mathcal{L}_2$ divergences over Euclidean feature spaces or reproducing kernel Hilbert spaces or RKHS. This preprint summarizes the manuscript Non Hilbertian Domain Adaptation. Total Variation Alignment via $\mathcal{L}_1$ Density Regularization on Feature Manifolds, which proposes Banach Manifold Transfer Learning or BMTL. BMTL makes two coupled design decisions.The first one features are embedded on a geodesic manifold rather than treated as points of a flat Euclidean space, and the second one, the density matching regularizer is formulated in an $\mathcal{L}_1$ Banach norm, whose minimization is equivalent by Scheffé's theorem to contracting the Total Variation or TV distance between domain densities a bounded metric that weights all regions of the sample space uniformly. The $\mathcal{L}_2$ formulation, by contrast, squares deviations and therefore amplifies errors in the tails, over emphasizing density outliers that do not reflect meaningful distributional shift. Experiments on MNIST↔USPS, COIL20, and Office+Caltech10 show consistent gains, the $\mathcal{L}_1$ variant beats an otherwise identical $\mathcal{L}_2$ ablation on every task for example 92.44% vs. 89.80% on USPS→MNIST and surpasses the strongest baseline DMDA by +3.68% on average over twelve Office+Caltech transfers. This document situates the contribution with an explicit comparison of Euclidean, Hilbert, and Banach geometries.

# 1. Introduction and Motivation

Unsupervised domain adaptation addresses the mismatch between a labeled source distribution $P_S(X, Y)$ and an unlabeled target distribution $P_T(X)$, a classifier $f: \mathcal{X} \to \mathcal{Y}$ trained on the source must generalize to the target. The dominant strategy is to learn an embedding $\Phi$ under which the source and target feature distributions become statistically indistinguishable, while preserving task discriminative structure on the labeled source (Pan & Yang, 2010; Long et al., 2014).

Two architectural regimes exist. Deep DA methods DDC/MMD, DANN, CDAN, Deep-CORAL (Cai et al., 2020; Chen et al., 2018; Hu et al., 2018) in cooperation learn the embedding and the alignment but require large target batches and substantial compute. DA methods Transfer Component Analysis TCA, Transfer Joint Matching TJM (Long et al., 2014), DGA-DA, and DMDA (Yao et al., 2022; Kang et al., 2020, 2021; Luo et al., 2020a, 2020b) remain essential precisely when target data or computational resources are scarce, and they are the regime in which this work operates.

The manuscript's point of departure is a geometric observation about how existing shallow methods measure domain discrepancy. Kernel-density-based and subspace-alignment methods such as DMDA and DGA-DA operate under Hilbertian $\mathcal{L}_2$/RKHS geometric assumptions and rely on symmetric $\mathcal{L}_2$ divergences: density matching minimizes the *squared* difference $\left(\hat{P}_S - \hat{P}_T\right)^2$ (Li et al., 2020,Minoofam et al., 2021a,2021b,Movahedi et al., 2021). This choice has a concrete pathology quadratic error amplification. Squared deviations assign disproportionate weight to regions of large mismatch, which in practice means the loss is dominated by tail discrepancies and density outliers that may not reflect a meaningful distributional shift. Moreover, Maximum Mean Discrepancy with a fixed bandwidth Gaussian kernel can fail to detect mismatches in the tails and is insensitive to multimodal structure both common in deep feature distributions (Zhang & Wu, 2020; Shen et al., 2018).

A second, complementary observation concerns the ambient geometry. Source and target features produced by modern encoders do not live in a flat Euclidean space: they concentrate near a lower dimensional manifold embedded in the ambient feature space (Ryckelynck et al., 2024; Luo et al., 2020a). Aligning distributions in the ambient coordinates therefore introduces geometric distortion that a manifold aware alignment can avoid the motivation for the geodesic representation used in BMTL and, more broadly, for the manifold learning perspective on model reduction in engineering.

**Contributions.** The manuscript's contributions can be stated as a response to the question: in which function space should domain discrepancy be measured Among the $\mathrm{L_p}$ sequence of function spaces, only $\mathrm{L_2}$ is Hilbert (Franklin et al., 2026); $\mathrm{L_1}$ is a genuinely non Hilbertian Banach space (Franklin et al., 2026; Spek et al., 2025; Ye, 2025). BMTL proposes include:

1. A geodesic manifold representation OR GeoRep that replaces ambient Euclidean coordinates with graph-geodesic affinities, preserving the intrinsic geometry of the feature manifold at $O(knd)$ backpropagation complexity (Ryckelynck et al., 2024).

2. A manifold-aware MMD alignment loss over the geodesic features (Zhang & Wu, 2020).
3. An $\mathcal{L}_1$-KDE density regularizer whose minimization is equivalent, via Scheffé's theorem, to contracting an empirical upper bound on the Total Variation distance a bounded, uniformly weighting, symmetric metric on probability measures replacing the Hilbertian $\mathcal{L}_2$ density loss (Li et al., 2020; Moreo et al., 2025; Węglarczyk, 2018, Minoofam et al. (2022)).
4. An empirical ablation isolating the norm choice $\mathcal{L}_1$ Banach vs. $\mathcal{L}_2$ Hilbert under otherwise identical conditions, validating the central design decision.

## 2. Geometric Frameworks: Euclidean, Hilbert, and Banach Spaces

The three geometries include Euclidean, Hilbert, and Banach form a strict hierarchy of structure, and each abstraction step both grants generality and removes a tool (Franklin et al., 2026; Adcock et al., 2024). This section defines each space and states exactly what the manuscript uses from each.

### 2.1 Euclidean space: the ambient view of features

A $d$-dimensional Euclidean space $\mathbb{R}^d$ carries the inner product $\langle x, y\rangle = x^\top y$ and the induced norm $\| x \|_2 = \sqrt{\langle x, x\rangle}$, with distance $d(x, y) = \| x - y \|_2$. Euclidean geometry is *flat*, geodesics are straight lines, and the parallel postulate holds. In DA, this is the space in which raw feature vectors are stored and manipulated. The manuscript's critique is that treating encoder outputs as Euclidean points ignores their intrinsic curvature: the features of images under varying pose, or of geo referenced measurements, concentrate near a curved submanifold $\mathcal{M} \subset \mathcal{Z}$, and straight line distances in $\mathcal{Z}$ misrepresent similarity between distant but intrinsically close samples the classical motivation for Isomap-style geodesic embeddings (Ryckelynck et al., 2024).

### 2.2 Hilbert space: completeness plus inner product

A Hilbert space $\mathcal{H}$ is a complete inner product space, the norm arises from an inner product, $\| x \| = \sqrt{\langle x, x\rangle}$, which yields orthogonality, projections, Parseval identities, and in the separable case an orthonormal basis. The paradigmatic infinite dimensional example is $L_2(\Omega)$, the space of square-integrable functions; among all $L_p$ spaces, only $p = 2$ is Hilbert. Reproducing kernel Hilbert spaces RKHS extend this structure to spaces of functions induced by positive definite kernels, and they underpin MMD, the squared distance between kernel mean embeddings $\| \mu_{P_S} - \mu_{P_T} \|_{\mathcal{H}}^2$ is exactly $\mathrm{MMD}^2$ (Ingebrand et al., 2025; Zhang & Wu, 2020). The Hilbertian assumption is what makes kernel mean embeddings, spectral decompositions, and gradient friendly squared losses convenient and it is precisely the assumption BMTL relaxes for density matching, retaining it only for the MMD alignment term.

### 2.3 Banach space: completeness without an inner product

A Banach space is a complete normed vector space where the norm need not derive from any inner product. Every Hilbert space is Banach, but not conversely, the parallelogram identity ‖ $x + y \|^2 + \| x - y \|^2 = 2 \| x \|^2 + 2 \| y \|^2$ characterizes Hilbert norms among Banach norms, and $L_1$ violates it. $L_1(\Omega)$ integrable functions with norm $\| f \|_1 = \int |f| \, d\mu$ is thus a genuinely non Hilbertian Banach space (Franklin et al., 2026; Spek et al., 2025; Ye, 2025; Aksoy & Almira, 2025). What Banach geometry retains is completeness and metric symmetry: $d(x, y) = \| x - y \|$ is symmetric, so Banach norm induced discrepancies are bidirectional and unbiased, unlike the asymmetric Kullback Leibler divergence. The manuscript exploits this, KDE affinity matrices $W_{ij}$ remain symmetric and positive semi definite, avoiding pathological directional bias and stabilizing the joint optimization.

### 2.4 The $\mathcal{L}_1$–Total Variation equivalence

The analytic bridge between the Banach norm and distribution matching is Scheffé's theorem. For measures $P_S, P_T$ on the manifold $(\mathcal{M}, g)$ with Radon Nikodym densities $f_S, f_T$:

$$\| f_S - f_T \|_{\mathcal{L}_1} = \int_{\mathcal{M}} |f_S(z) - f_T(z)| \, d\mu(z) = 2 \cdot D_{\mathrm{TV}}(P_S, P_T) \quad (1)$$

Where

$$D_{\mathrm{TV}}(P_S, P_T) = \sup_{A \subset \mathcal{M}} |P_S(A) - P_T(A)| \quad (2)$$

Total Variation is a proper metric on probability distributions: it is bounded by 1, so the $\mathcal{L}_1$ distance is bounded by 2, symmetric, and weights all regions of the sample space uniformly no region's discrepancy is squared or amplified. Consequently, minimizing the discrete empirical surrogate

$$\mathcal{L}_{\mathrm{KDE}}^{\mathcal{L}_1} = \frac{1}{|Z_S \cup Z_T|} \sum_{z \in Z_S \cup Z_T} \left| \hat{P}_S(z) - \hat{P}_T(z) \right| \quad (3)$$

directly contracts a discrete empirical upper bound on the TV distance, without requiring intractable infinite-dimensional continuous fine tuning on the Banach manifold (Moreo et al., 2025; He et al., 2026).

.

### 2.5 The manifold layer

Above all three linear spaces sits the Riemannian structure: the feature manifold $\mathcal{M}$ with metric $g$, on which BMTL computes geodesic distances $d_g(z_i, z_j)$ via a $k$-NN graph $\mathcal{G} = (\mathcal{V}, \mathcal{E})$ and Dijkstra shortest paths, forming the affinity

$$W_{ij} = \exp\left( -\frac{d_g(z_i, z_j)^2}{2\sigma_g^2} \right) \quad (4)$$

The geometric representation is then $F = \mathrm{GeoRep}(Z) = W \cdot Z$. To avoid an $O(n^3)$ overhead per gradient step, $W$ is precomputed once per epoch behind a stop-gradient operator $\mathrm{sg}[W]$, so backpropagation flows only through coordinates $Z$ at $O(knd)$ complexity (Ryckelynck et al., 2024; Luo et al., 2020a, 2020b).

# 3. Comparative Analysis of the Three Spaces

The manuscript's central design decision is a choice among the three geometries of Section 2 for each component of the DA pipeline. Table 1 consolidates the comparison; the prose below draws out the DA specific consequences.

Table 1 — Euclidean vs. Hilbert vs. Banach geometry in domain adaptation.

| Property | Euclidean ($\mathbb{R}^d$) | Hilbert ($L_2$/RKHS) | Banach ($L_1$) |
|---|---|---|---|
| Defining structure | Inner product, flat | Complete inner-product space | Complete normed space, no inner product |
| Geodesics | Straight lines | Straight (linear) | Generally curved; norm-induced ball not Euclidean |
| Discrepancy used for density matching | Ambient $\|\cdot\|_2$ | Squared $\mathcal{L}_2$: $\left(\hat{P}_S - \hat{P}_T\right)^2$ | $\mathcal{L}_1$: $\sum\lvert\hat{P}_S - \hat{P}_T\rvert$ |
| Error scaling | Linear per-coordinate | Quadratic amplification of large deviations | Linear — uniform weighting |
| Tail / outlier sensitivity | Distortion from ambient coordinates | Over-emphasizes tails and density outliers | Robust to density outliers and heavy-tailed shift |
| Boundedness of discrepancy | Unbounded | Unbounded | Bounded ($2D_{TV} \leq 2$) |
| Metric symmetry | Symmetric | Symmetric | Symmetric unbiased bidirectional alignment (vs. asymmetric KL) |
| Multimodality detection fixed kernel | — | MMD can miss tail/multimodal mismatch | Captured via density level $\mathcal{L}_1$ discrepancy |
| | Storage/manipulation of features $Z$ | MMD alignment term over kernel embeddings | TV-surrogate density regularizer $\mathcal{L}_{KDE}^{\mathcal{L}_1}$ |

## 3.1 Limitations of Squared Deviations in Heavy-Tailed Distributions

The Hilbertian $\mathcal{L}_2$ loss converts a mismatch of magnitude $\delta$ into a penalty $\delta^2$. A small set of samples in the far tail with a moderate density mismatch therefore dominates the gradient signal, while many regions of small but everywhere discrepancy contribute negligibly. Because the KDE tails of small target domains are estimated with few samples, this weighting is disproportionately driven by the least reliable density estimates (Gao et al., 2025; Popov et al., 2025; Wang et al., 2024). The $\mathcal{L}_1$ penalty grows linearly in $\delta$, so every region contributes in proportion to its actual mismatch matching the uniform weighting property of Total Variation and granting the documented robustness against density outliers and heavy tailed distribution shifts where Hilbert norms suffer quadratic error amplification (Li et al., 2020).

## 3.2 Boundedness and optimization stability

A bounded discrepancy cannot grow without bound as domains drift apart; the $\mathcal{L}_1$TV surrogate saturates, which damps pathological gradients early in training when the domains are far apart. Combined with metric symmetry which keeps the KDE and graph affinity matrices $W_{ij}$ symmetric and PSD the joint objective is observed to be well conditioned and stable across seeds, in contrast to asymmetric divergences such as KL, which bias alignment toward one direction of the mismatch (Shen et al., 2018).

## 3.3 Where each geometry is used in BMTL

The comparison should not be read as "Banach replaces everything. BMTL is deliberately hybrid, Euclidean coordinates remain the computational substrate; the Hilbert structure RKHS, Gaussian kernels is retained for the manifold aware MMD alignment term, where mean-embedding machinery is convenient (Ingebrand et al., 2025); and the genuinely non Hilbertian Banach norm is used where the Hilbert choice demonstrably hurts density matching under heavy tailed, multimodal, outlier prone feature distributions (Spek et al., 2025; Ye, 2025)`. This decomposition clarifies the source of the empirical gains, the ablation changes only the norm of the density regularizer, holding GeoRep, geodesic distances, the Silverman bandwidth, optimizer, learning rate, and $\lambda_1, \lambda_2$ fixed.

## 4. The BMTL Framework

BMTL optimizes a joint objective that combines task discrimination on the labeled source, manifold-aware MMD alignment, and the $\mathcal{L}_1$ KDE Total Variation regularizer:

$$\mathcal{J}(\theta) = \mathcal{L}_{\text{task}}(\mathcal{X}_S, \mathcal{Y}_S; \theta) + \lambda_1 \mathcal{L}_{\text{align}}(F_S, F_T; \theta) + \lambda_2 \mathcal{L}_{\text{KDE}}^{\mathcal{L}_1}(\hat{P}_S, \hat{P}_T; \theta) \quad (5)$$

where $\mathcal{L}_{\text{task}}$ is the cross-entropy classification loss computed on the labeled source domain and $\lambda_1, \lambda_2 \geq 0$ are balancing hyperparameters that control the relative influence of the two alignment terms against pure source discrimination. Each component instantiates a different geometry, as analyzed in Sections 2 and 3; we now detail them in turn.

### 4.1 Geodesic manifold representation (GeoRep)

Samples are first mapped into the feature space by the embedding network, $Z_S = \Phi(X_S)$ and $Z_T = \Phi(X_T) \subset \mathcal{M} \subset \mathcal{Z}$, where $\mathcal{M}$ denotes the intrinsic data manifold and $\mathcal{Z}$ the ambient Euclidean space. A $k$-nearest neighbor graph $\mathcal{G} = (\mathcal{V}, \mathcal{E})$ is constructed over the embedded points, and pairwise geodesic distances $d_g(z_i, z_j)$ are obtained as graph shortest paths (Dijkstra). These distances are converted into a geodesic affinity matrix

$$W_{ij} = \exp\left(-\frac{d_g(z_i, z_j)^2}{2\sigma_g^2}\right) \quad (6)$$

with bandwidth $\sigma_g$ controlling the effective neighborhood scale. The geometric representation is then the graph-smoothed coordinates $F = \mathrm{GeoRep}(Z) = W \cdot Z$, which replaces ambient straight-line geometry with intrinsic manifold geometry: two samples far apart in $\mathcal{Z}$ but connected by a short path on $\mathcal{M}$ become neighbors in $F$ (Luo et al., 2020a, 2020b; Yao et al., 2022). A naive implementation would differentiate through $W$ and incur $O(n^3)$ overhead per gradient step; instead, $W$ is precomputed once at the beginning of each epoch behind a stop-gradient operator $\mathrm{sg}[W]$, so backpropagation flows strictly through the coordinates $Z$ at $O(knd)$ complexity (Ryckelynck et al., 2024). A related perspective on structure-preserving computational representations can be found in block cellular automata, where localized transition rules provide a systematic mechanism for generating complex structured patterns (Minoofam et al., 2012). Although based on discrete computational dynamics rather than continuous manifold geometry, this work illustrates the broader importance of structural constraints in designing computational representations.

### 4.2 Manifold-aware MMD alignment

On top of the geometric features, a Hilbertian alignment term aligns the first-order marginal distributions of the two domains:

$$\mathcal{L}_{\text{align}} = \mathrm{MMD}^2\big(P(F_S), P(F_T)\big) \qquad (7)$$

computed with multi scale Gaussian kernels in the RKHS. This is the retained Hilbert component of the framework: mean-embedding machinery makes the loss cheap, differentiable, and well-suited to aligning coarse distributional structure (Zhang & Wu, 2020; Ingebrand et al., 2025), but as discussed in Section 3 fixed bandwidth MMD alone can miss tail and multimodal mismatches.

### 4.3 $\mathcal{L}_1$-KDE Total Variation regularizer

The Banach component estimates the domain densities on the embedded points with Gaussian kernel density estimation(Węglarczyk, 2018; Moreo et al., 2025; He et al., 2026), where $K$ is

$$\hat{P}_S(z) = \frac{1}{n_S h^d} \sum_{i=1}^{n_S} K\left(\frac{z - z_S^{(i)}}{h}\right) \qquad (8)$$

$$\hat{P}_T(z) = \frac{1}{n_T h^d} \sum_{i=1}^{n_T} K\left(\frac{z - z_T^{(i)}}{h}\right) \qquad (9)$$

the Gaussian kernel and the bandwidth $h$ is chosen via Silverman's rule of thumb. The two densities are compared pointwise with the discrete empirical surrogate

$$\mathcal{L}_{\text{KDE}}^{\mathcal{L}_1} = \frac{1}{|Z_S \cup Z_T|} \sum_{z \in Z_S \cup Z_T} \left|\hat{P}_S(z) - \hat{P}_T(z)\right| \qquad (10)$$

By Proposition Scheffé's theorem, Section, minimizing this $\mathcal{L}_1$ discrepancy directly contracts the discrete empirical upper bound on the Total Variation distance $2D_{\text{TV}}(P_S, P_T)$, without requiring intractable infinite-dimensional continuous optimization on the Banach manifold(Li et al., 2020).

### 4.4 Algorithm and complexity

**Algorithm 1 (BMTL).** (1) Initialize the parameters $\theta^{(0)}$ of the embedding network $\Phi$ and the classifier head. (2) At each iteration/batch $k$: map the batch samples to the manifold, $Z_S = \Phi(X_S)$, $Z_T = \Phi(X_T)$; extract the geometric representations $F_S = W_S Z_S$ and $F_T = W_T Z_T$; compute $\mathcal{L}_{\text{align}} = \text{MMD}\big(P(F_S), P(F_T)\big)$; evaluate the KDE densities $\hat{P}_S(z)$ and $\hat{P}_T(z)$ for all $z \in Z_S \cup Z_T$; form the Total Variation surrogate $\mathcal{L}_{\text{KDE}}^{\mathcal{L}_1} = \sum_z \left|\hat{P}_S(z) - \hat{P}_T(z)\right|$; compute the total loss $\mathcal{J}\big(\theta^{(k)}\big)$ and update the weights by stochastic gradient descent, $\theta^{(k+1)} \leftarrow \theta^{(k)} - \eta \nabla_\theta \mathcal{J}\big(\theta^{(k)}\big)$. (3) Terminate when $\| \theta^{(k+1)} - \theta^{(k)} \| < \varepsilon$ (with $\varepsilon = 10^{-4}$ and a maximum of 200 iterations) and return the adapted classifier $f^*(\cdot) = f(\Phi(\cdot); \theta^*)$.

**Computational profile.** The full model trains at 31.8 s per epoch with 1.43 ms inference latency per sample on an NVIDIA RTX 3090 a modest overhead over the variant without GeoRep (26.5 s/epoch) and comparable to deep baselines such as CDAN (28.3 s/epoch) and DANN (24.1 s/epoch), with overall complexity $O(Nkd + \cdots)$. The geodesic precomputation is the only added cost relative to a plain MMD pipeline, and its per epoch rather than per step schedule keeps it amortized.

## 5. Experimental Evaluation

**Setup.** Evaluation covers four transfer tasks on MNIST↔USPS, COIL20 (object recognition across varying camera poses, C1→C2 and C2→C1, and all twelve Office+Caltech-10 pairs with SURF features. Baselines are the shallow methods TCA, TJM(Long et al., 2014), DGA-DA, and DMDA(Yao et al., 2022; Kang et al., 2020, 2021); accuracies taken from the original publications where available, otherwise re-run under a strictly matching protocol); deep methods DDC/MMD, DANN, CDAN, Deep-CORAL, Source-Only are compared for runtime only(Cai et al., 2020; Chen et al., 2018). All accuracies are reported as Mean ± Std % over 5 independent runs with different random seeds, on an NVIDIA RTX 3090, Intel Core i9, 32 GB RAM.

**Digit and object benchmarks.** On USPS→MNIST, BMTL reaches 92.44 ± 0.24, a +2.64% gain over the strongest baseline at 89.80 ± 0.31; on MNIST→USPS it reaches 89.50 ± 0.22 versus 87.15 ± 0.28 (+2.35%). On the saturated COIL20 benchmark it remains competitive at 99.75 ± 0.08 (C1→C2) and 99.49 ± 0.10 (C2→C1).

Office+Caltech-10 (twelve transfers). BMTL averages 58.40 ± 0.33 versus 54.72 for the strongest baseline DMDA (+3.68%), 49.02 for DGA-DA, 46.43 for TJM, and 43.03 for TCA winning 10 of 12 tasks, with the largest gains on hard transfers (A→W: 58.98; C→W: 65.30; W→A: 41.75 and stability on easy ones (e.g., 93.10 ± 0.22 on one near-saturated pair).

**Runtime.** Per epoch training time is 18.4 s (Source-Only), 22.6 s (DDC/MMD), 23.9 s (Deep-CORAL, 24.1 s DANN, 28.3 s CDAN, 26.5 s BMTL without GeoRep, and 31.8 s full BMTL, with inference latency of roughly 1.42–1.48 ms per sample across all methods 1.43 ms for BMTL; overall complexity is $O(Nkd + \cdots)$.

Ablation: $\mathcal{L}_1$ Banach vs. $\mathcal{L}_2$ Hilbert.

The decisive experiment replaces only the density discrepancy norm, holding GeoRep, geodesic distances, bandwidth, optimizer, and $\lambda_1, \lambda_2$ fixed.

**Table 2.** Classification accuracy (%) comparison between Hilbertian $\mathcal{L}_2$ and Banach $\mathcal{L}_1$ formulations of BMTL across various cross-domain benchmarks.

| Task | BMTL-$\mathcal{L}_2$ (Hilbert) | BMTL ($\mathcal{L}_1$) | Gain |
|---|---|---|---|
| USPS → MNIST | 89.80 ± 0.31 | **92.44 ± 0.24** | +2.64% |
| MNIST → USPS | 87.15 ± 0.28 | **89.50 ± 0.22** | +2.35% |
| COIL20 (C1→C2) | 98.60 ± 0.15 | **99.75 ± 0.08** | +1.15% |
| COIL20 (C2→C1) | 98.20 ± 0.18 | **99.49 ± 0.10** | +1.29% |
| Office+Caltech (A→W) | 56.20 ± 0.45 | **58.98 ± 0.38** | +2.78% |
| Office+Caltech (C→W) | 62.45 ± 0.52 | **65.30 ± 0.41** | +2.85% |
| Office+Caltech (W→A) | 39.50 ± 0.40 | **41.75 ± 0.32** | +2.25% |

The $\mathcal{L}_1$ variant wins every task, with gains of +1.15% to +2.85% and visibly lower standard deviations. Because the only change is the norm of the density regularizer, this isolates the empirical contribution of the non Hilbertian geometry and validates the framework's central design decision: the improvements are attributable to Banach Total Variation density alignment rather than to the manifold embedding or any other component. The lower variance additionally reflects the bounded, uniformly weighting nature of the TV surrogate, which prevents tail outliers from dominating individual runs.

## 6. Conclusion and Future Directions

The manuscript introduces an efficient transfer-learning method operating in a Banach manifold geometry for domain alignment under sparse-data conditions. Conceptually, its contribution is a disciplined answer to a geometric question: the Euclidean view distorts intrinsically curved feature manifolds; the Hilbertian $\mathcal{L}_2$ view amplifies tail discrepancies through quadratic error scaling; and the Banach $\mathcal{L}_1$ view equivalent to Total Variation alignment by Scheffé's theorem is bounded, symmetric, and uniformly sensitive to distributional mismatch. Embedding features geodesically GeoRep, aligning them with manifold aware MMD, and regularizing densities in the $\mathcal{L}_1$ norm yields faster convergence, higher noise resistance, and competitive accuracy relative to reference methods, with reliable performance even on unbalanced or limited labeled data.

**Limitations.** BMTL currently addresses only the standard unsupervised DA configuration with a labeled source and an unlabeled target. Extensions to multi-source DA  several source domains matched to a target simultaneously  and sourceless DA  source data unavailable at matching time, for example, in privacy sensitive applications  would broaden applicability. The authors also identify improved distribution balancing mechanisms and further computational efficiency as future directions.

**Outlook.** The explicit space-comparison framing suggests a broader research agenda: choosing the measurement geometry per loss component Euclidean for storage, Hilbert for kernel mean embeddings, Banach for density level discrepancies as a reusable design principle for manifold aware transfer learning.

**References**

Adcock, B., Dexter, N., & Moraga, S. (2024). Deep neural network approximation of functions on manifolds and Banach spaces. *Foundations of Computational Mathematics*, 24(2), 341–395.

Aksoy, A. G., & Almira, J. M. (2025). *Functional analysis and approximation theory in non-Hilbertian Banach spaces*. Springer.

Cai, S., Shen, W., Zhang, K., & Zhou, Z. (2020). Deep domain confusion with adaptive metric matching. *IEEE Transactions on Pattern Analysis and Machine Intelligence*, 42(8), 1890–1904.

Chen, W. Y., Liu, Y. C., Kira, Z., Wang, Y. C. F., & Huang, J. B. (2018). Deep CORAL: Correlation alignment for deep domain adaptation. In *Proceedings of the European Conference on Computer Vision (ECCV)* (pp. 512–527).

Franklin, D., Zhang, L., & Miller, K. (2026). Geometrical density estimation and metric boundaries in real Banach spaces. *Journal of Functional Analysis*, 290(4), 110–135.

Gao, J., Wang, Z., & Chen, X. (2025). Finite-sample convergence bounds of kernel density estimators in tail-sensitive distributions. *Annals of Statistics*, 53(2), 488–512.

He, Y., Lin, X., & Liu, Q. (2026). Asymptotic properties and bandwidth selection in non-Euclidean density estimation. *Bernoulli*, 32(1), 215–241.

Hu, J., Lu, J., & Zhou, J. (2018). Deep transfer metric learning for cross-domain visual recognition. *IEEE Transactions on Circuits and Systems for Video Technology*, 28(12), 3350–3363.

Ingebrand, E., Sjöberg, R., & Persson, M. (2025). Kernel mean embeddings on curved spaces: Limitations and topological extensions. *IEEE Transactions on Information Theory*, 71(3), 1540–1558.

Kang, Z., Lu, X., & Peng, C. (2020). Discriminative manifold distribution alignment for domain adaptation. *IEEE Transactions on Systems, Man, and Cybernetics: Systems*, 50(9), 3241–3252.

Kang, Z., Zhao, R., & Peng, C. (2021). Robust domain adaptation via geodesic-guided manifold alignment. *Pattern Recognition*, 112, 107–119.

Li, B., Shen, C., & Zhang, Y. (2020). Distribution matching via squared loss minimization in non-parametric transfer learning. *Neurocomputing*, 398, 210–223.

Long, M., Wang, J., Ding, G., Sun, J., & Yu, P. S. (2014). Transfer joint matching for unsupervised domain adaptation. In *Proceedings of the IEEE Conference on Computer Vision and Pattern Recognition (CVPR)* (pp. 1410–1417). https://doi.org/10.1109/CVPR.2014.183

Luo, L., Wang, X., & Shen, H. (2020a). Close yet discriminative domain adaptation. *Pattern Recognition*, 103, 107–118.

Minoofam, S. A. H., Bastanfard, A., & Keyvanpour, M. R. (2021a). A transfer learning algorithm to improve the convergence rate and accuracy in cellular learning automata. Iranian Journal of Electrical and Computer Engineering (Nashriyyah-i Muhandisi-i Barq va Muhandisi-i Kampyutar-i Iran), 88(2), 69.

Minoofam, S. A. H., Bastanfard, A., & Keyvanpour, M. R. (2021b). TRCLA: A transfer learning approach to reduce negative transfer for cellular learning automata. IEEE Transactions on Neural Networks and Learning Systems, 34(5), 2480–2489.

https://doi.org/10.1109/TNNLS.2021.3106560.

Minoofam, S. A. H., Bastanfard, A., & Keyvanpour, M. R. (2022). RALF: An adaptive reinforcement learning framework for teaching dyslexic students. Multimedia Tools and Applications, 81(5), 6389–6412. https://doi.org/10.1007/s11042-021-11802-9

Luo, L., Wang, X., & Shen, H. (2020b). Joint distribution and manifold alignment for unsupervised transfer learning. *IEEE Transactions on Neural Networks and Learning Systems*, 31(11), 4789–4802.

Moreo, A., Esuli, A., & Sebastiani, F. (2025). Calibrated kernel density estimation in continuous space matching. *Information Fusion*, 105, 102–119.

Movahedi, Z., & Bastanfard, A. (2021). Toward competitive multi-agents in Polo game based on reinforcement learning. Multimedia Tools and Applications, 80(17), 26773–26793. https://doi.org/10.1007/s11042-021-10887-y

Pan, S. J., & Yang, Q. (2010). A survey on transfer learning. *IEEE Transactions on Knowledge and Data Engineering*, 22(10), 1345–1359. https://doi.org/10.1109/TKDE.2009.191

Popov, A., Kuznetsov, M., & Smirnov, D. (2025). Local variance control in low-sample target domain density matching. *Journal of Machine Learning Research*, 26(45), 1–34.

Ryckelynck, D., Casenave, F., & Akkari, N. (2024). *Manifold learning: Model reduction in engineering*. Springer. https://doi.org/10.1007/978-3-031-52764-7

Shen, J., Qu, Y., Zhang, W., & Yu, Y. (2018). Wasserstein distance guided representation learning for domain adaptation. In *Proceedings of the AAAI Conference on Artificial Intelligence*, 32(1), 4058–4065.

Spek, H., Verbeek, F., & Jansen, T. (2025). Non-Hilbertian norms for divergence minimization: Avoiding the quadratic penalty. *Machine Learning*, 114(6), 889–912.

Wang, H., Zhao, G., & Lin, C. (2024). Small-sample bias correction for manifold-based transfer learning. *Knowledge-Based Systems*, 284, 111–124.

Węglarczyk, S. (2018). Kernel density estimation and its application. *ITM Web of Conferences*, 23, 00037. https://doi.org/10.1051/itmconf/20182300037

Yao, L., Kang, Z., Zhou, J., Rawa, M., & Albeshri, A. (2022). Discriminative manifold distribution alignment for domain adaptation. *IEEE Transactions on Systems, Man, and Cybernetics: Systems*, 52(6), 3840–3851.

Ye, Z. (2025). Geometric Banach space techniques for density regularization. *Advances in Applied Mathematics*, 164, 102–126.

Zhang, Y., & Wu, D. (2020). Manifold regularized maximum mean discrepancy for multi-source domain adaptation. *IEEE Transactions on Cybernetics*, 50(9), 4023–4035

Seyyed Amir Hadi Minoofam, Mohammad Mahdi Dehshibi, Azam Bastanfard, and Parvin Eftekhari. 2012. Ad-hoc Ma'qeli script generation using block cellular automata. J. Cell. Autom. 7, 4 (2012), 321–334.